\documentclass{article}

\usepackage{amsmath}
\usepackage{microtype}
\usepackage{graphicx}
\usepackage{caption}
\usepackage{subcaption}
\usepackage{booktabs} % for professional tables
\usepackage{listings}
\usepackage{multirow}
\usepackage{xurl}

\usepackage{hyperref}

\usepackage[accepted]{icml2021}

\usepackage{color}

\usepackage{ulem}

\icmltitlerunning{Revisiting Language Encoding in Learning Multilingual Representations}

\begin{document}

\twocolumn[
\icmltitle{Revisiting Language Encoding in Learning Multilingual Representations}



\icmlsetsymbol{equal}{*}

\begin{icmlauthorlist}
\icmlauthor{Shengjie Luo}{equal,pku}
\icmlauthor{Kaiyuan Gao}{equal,hust}
\icmlauthor{Shuxin Zheng}{ms}
\icmlauthor{Guolin Ke}{ms}
\icmlauthor{Di He}{ms}
\icmlauthor{Liwei Wang}{pku}
\icmlauthor{Tie-Yan Liu}{ms}
\end{icmlauthorlist}

\icmlaffiliation{pku}{Peking University}
\icmlaffiliation{hust}{Huazhong University of Science and Technology}
\icmlaffiliation{ms}{Microsoft Research}

\icmlcorrespondingauthor{Shuxin Zheng}{shuxin.zheng@microsoft.com}
\icmlcorrespondingauthor{Di He}{dihe@microsoft.com}

\icmlkeywords{Language Projection, Multilingual}

\vskip 0.3in]



\printAffiliationsAndNotice{\icmlEqualContribution}  % leave blank if no need to mention equal contribution

\begin{abstract}
Transformer has demonstrated its great power to learn contextual word representations for multiple languages in a single model. To process multilingual sentences in the model, a learnable vector is usually assigned to each language, which is called ``language embedding''. The language embedding can be either added to the word embedding or attached at the beginning of the sentence. It serves as a language-specific signal for the Transformer to capture contextual representations across languages. In this paper, we revisit the use of language embedding and identify several problems in the existing formulations. By investigating the interaction between language embedding and word embedding in the self-attention module, we find that the current methods cannot reflect the language-specific word correlation well. Given these findings, we propose a new approach called Cross-lingual Language Projection (XLP) to replace language embedding. For a sentence, XLP projects the word embeddings into language-specific semantic space, and then the projected embeddings will be fed into the Transformer model to process with their language-specific meanings. In such a way, XLP achieves the purpose of appropriately encoding ``language'' in a multilingual Transformer model.  Experimental results show that XLP can freely and significantly boost the model performance on extensive multilingual benchmark datasets. Codes and models will be released at \url{https://github.com/lsj2408/XLP}.

\end{abstract}

\section{Introduction}

\begin{figure*}[ht]
\vskip 0.2in
\begin{center}
\centerline{\includegraphics[scale=0.9]{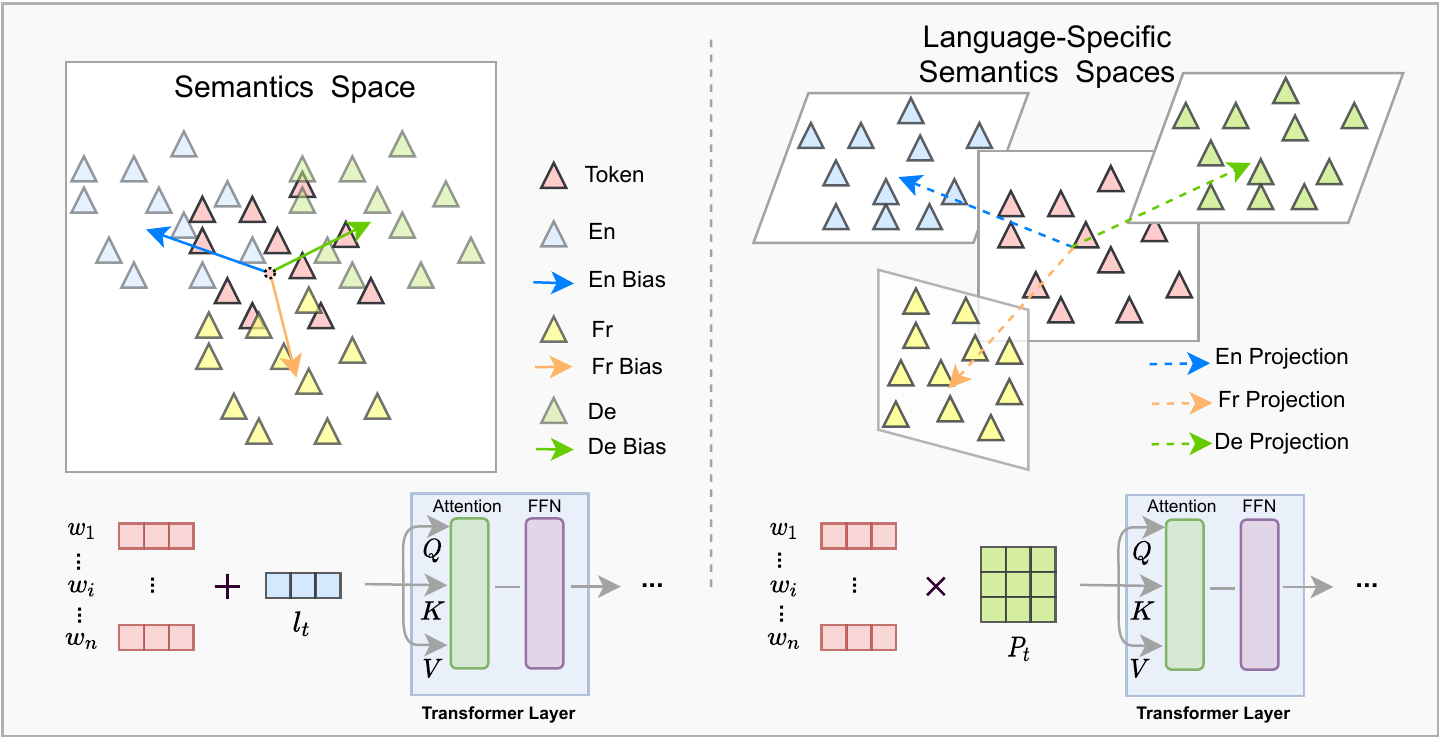}}
\caption{\textbf{An illustration of the language embedding approach (additive, left) and our proposed Cross-lingual Language Projection (XLP, right)}. For a sentence from language $t$, the language embedding $l_t$ is added to each word embedding and then fed into the Transformer layers. XLP projects word embeddings into language-specific semantic space using projection function $P_t$. Then the projected embeddings will be fed into the Transformer model to process with their language-specific meanings. In both figures, we omit the positional embedding for a better comparison of the two methods.}
\label{fig:xlp-illustration}
\end{center}
\vskip -0.2in
\end{figure*}

Learning a universal NLP model that supports multiple languages is usually a necessity in practical systems, such as multilingual machine translation, and multilingual sentence understanding ~\cite{dong-etal-2015-multi,zoph2016multisource,firat2016multiway,wu2016googles,johnson2017googles,lee2017fully,firat2016zeroresource,gu2018universal,lewis2020mlqa,lample2019crosslingual,conneau2020unsupervised}. Transformer~\cite{vaswani2017attention} is the most widely used neural network architecture in natural language processing. To handle sentences from different languages, people design ways to provide the ``language'' signal into the multilingual Transformer. Language embedding \cite{tan2019multilingual,lample2019crosslingual,huang2019unicoder,chi2019crosslingual,liu2020multilingual,tang2020multilingual} is a popular choice which views each language as a symbol with a learnable embedding vector. Previous works provide two approaches to using the language symbol: attaching it to the beginning of the sentence or adding its embedding to the word embedding at each position. Using such information, Transformers try to learn the word meanings in the corresponding language and obtain the contextual word representations accordingly. 

In this work, we revisit the use of language embedding in Transformer and find that the current approaches may be ineffective in learning multilingual representations. To show the problem clearly, we study the interaction between word embedding and language embedding in the Transformer layers and find that in both approaches, the word-language correlation will be computed in the self-attention module. We question whether this word-language correlation is useful in capturing semantic relations of words at different positions in the sentence. According to our empirical study, we observe that this correlation seems to reflect the popularity of a word appearing in a language to a certain extent. Obviously, such popularity cannot reflect whether two words have a strong semantic relationship in a language.

The analysis inspires us to think further about the proper way to encode ``language'' in a multilingual Transformer model. As the vocabulary is shared (i.e., one word-unit corresponds to one embedding vector), one word-unit may appear in multiple languages and have different meanings. We hope that with the language encoding, the Transformer model can receive language-specific word meaning and learn the contextual word representation based on that. Motivated by this, we propose a novel language encoding called Cross-lingual Language Projection (XLP). Instead of viewing ``languages'' as vectors, we view different ``languages'' as different projection functions, e.g., linear transformation matrice with learnable parameters. Given any sentence, XLP projects the word embeddings into language-specific semantic space using the corresponding projection function. Then the Transformer takes the projected word embeddings as input, calculates word-word correlations in the language-specific semantic space, and obtains the representation of the words and the sentence. See Figure \ref{fig:xlp-illustration} for an illustration.

XLP is conceptually simple and easy to implement with barely additional computation overhead but shows promising performance gain on a wide range of multilingual applications. To be specific, XLP gains 1.2\% of accuracy improvement on the zero-shot cross-lingual task XNLI \cite{conneau2018xnli} comparing to the previous state-of-the-art pre-trained model. XLP also shows better performance on multilingual translation tasks. It consistently improves the BLEU scores on translating different languages to English compared to the baseline models. 

\section{Preliminary}

\subsection{Attention Module}

The attention module \citep{vaswani2017attention} is formulated as querying a dictionary with key-value pairs, e.g., $\text{Attention}(Q,K,V)=\text{softmax}(\frac{QK^T}{\sqrt{d}})V$, where $d$ is the dimensionality of the hidden representations, and $Q$ (Query), $K$ (Key), $V$ (Value) are specified as the hidden representations of the previous layer. The multi-head variant of the attention module is popularly used which allows the model to jointly attend to the information from different representation sub-spaces, and is defined as
\begin{align}
\text{Multi-head}(Q,K,V) ={}& \text{Concat} (\text{head}_1,\cdots,\text{head}_H)W^O \nonumber\\
\text{head}_k ={}& \text{Attention}(QW_k^Q, KW_k^K,VW_k^V), \nonumber
\end{align}
where $W_k^Q\in \mathcal{R}^{d \times d_K}$, $W_k^K\in \mathcal{R}^{d \times d_K}$, $W_k^V\in \mathcal{R}^{d\times d_V}$, and $W^O\in \mathcal{R}^{H d_V \times d}$ are learnable project matrices, $H$ is the number of heads. $d_K$ and $d_V$ are the dimensionalities of Key and Value. 

The self-attention module is one of the key components in Transformer and BERT encoder \citep{devlin2019bert}. For simplicity, we use the single-head self-attention module and set $d_K = d_V = d$ for a demonstration. We denote $x^l=(x^l_{1}, x^l_{2} \cdots, x^l_{n})$
as the input to the self-attention module in the $l$-th layer, where $n$ is the length of the sequence and each vector $x^l_i \in \mathcal{R}^d$ is the contextual representation of the token at position $i$. $z^l=(z^l_{1}, z^l_{2} \cdots, z^l_{n})$ is the output of the attention module. Then, the self-attention module can be written as
\begin{eqnarray}\label{eqn:general-att2}
z^l_i=\sum_{j=1}^n \frac{\exp (\alpha_{ij})}{\sum_{j'=1}^n\exp (\alpha_{ij'})}(x^l_jW^{V,l}), \\
\text{where } \alpha_{ij}=\frac{1}{\sqrt{d}} (x^l_iW^{Q,l})(x^l_jW^{K,l})^T.
\label{eqn:self-attn}
\end{eqnarray}
As we can see, in any sentence, the self-attention module calculates the correlation between information at different positions, and use the correlation (i.e., attention) to obtain the contextual representation of each word by considering its surroundings. 

\subsection{Language Encoding in Multilingual Transformer}
When training a multilingual model, a shared vocabulary (of words or sub-words) covering all the languages is firstly prepared. A learnable word embedding is assigned to each word in the vocabulary. Then a Transformer model, which takes the word embeddings (and the positional embeddings) as input, is optimized using pre-defined objective functions on the multilingual training data. For example, in multilingual machine translation, an encoder-decoder Transformer is trained to maximize the conditional log-likelihood of the target sentence given the source sentence using translation pairs in all the languages.

However, we usually need to feed the model with an additional signal in terms of which language the sentence comes from. Sometimes, such information is essential. For example, in multilingual machine translation, the model can generate the translation results only if we provide the name of the target language we are requesting. To encode such information, previous works design a specific symbol for each language with a learnable embedding vector. There are generally two approaches to use language embeddings. The first approach (attaching approach for short) attaches the corresponding symbol to the beginning of the sentence \citep{wu2016googles,johnson2017googles,liu2020multilingual,tang2020multilingual}. The second approach (additive approach for short) adds the language embedding to the word embedding at each position \citep{tan2019multilingual,lample2019crosslingual,huang2019unicoder,chi2019crosslingual}. With the help of language embeddings, the model receives the ``language'' information explicitly from the input and learns the sentence representations through Transformer layers.

%%%%%%%%%%%%%%%%%%%%%%%%%%%%%%%%%%%%%%%%%%%%%
%        Modification Begin
%%%%%%%%%%%%%%%%%%%%%%%%%%%%%%%%%%%%%%%%%%%%%

\section{Cross-Lingual Language Projection}

\begin{figure}[ht]
\vskip 0.2in
\begin{center}
\centerline{\includegraphics[width=0.5\textwidth]{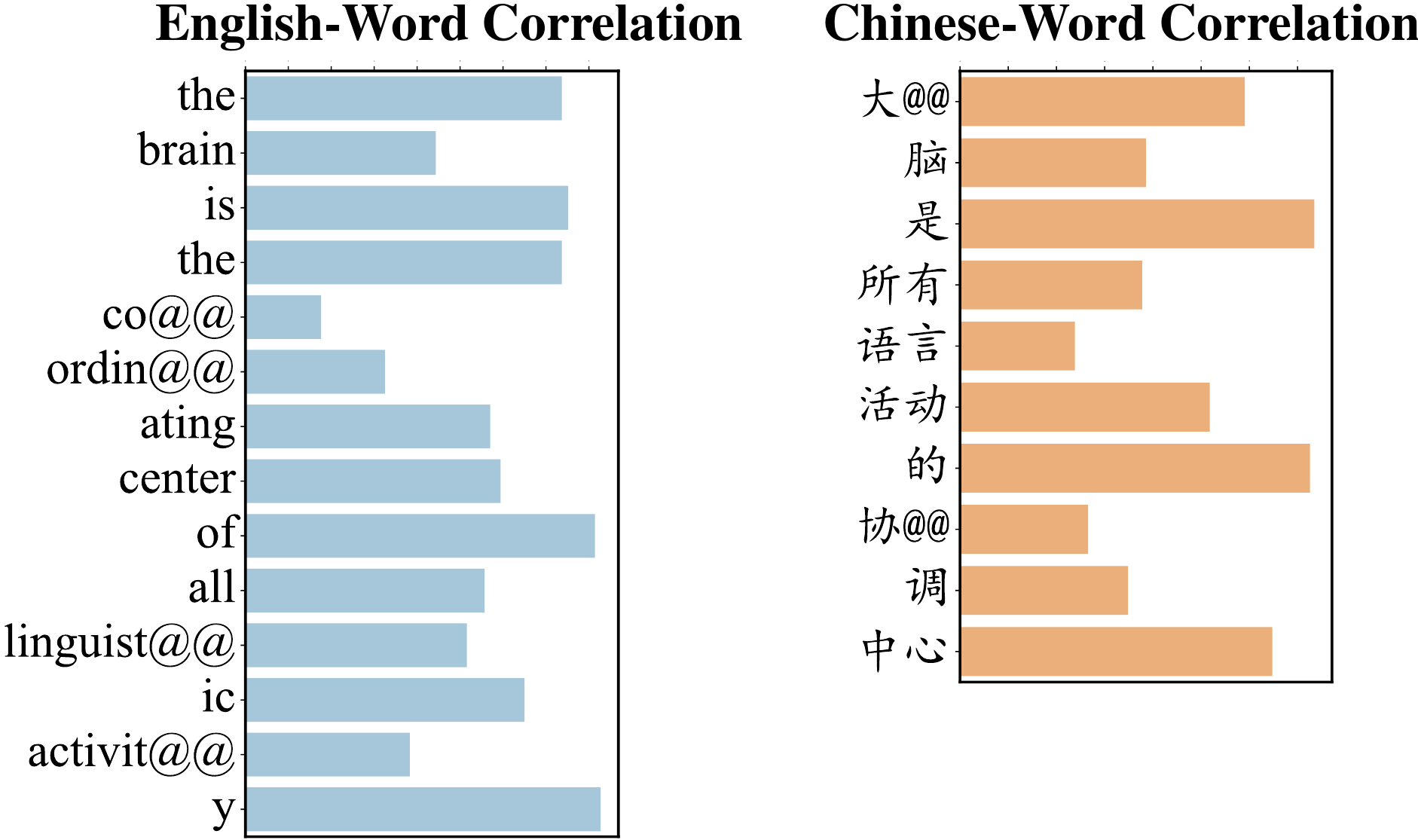}}
\caption{\textbf{Visualizations of the word-language correlation in Eq.\eqref{eqn:expand-attn}.} We use the publicly released pre-trained XLM model and calculate the correlations on one head. We sampled two synonymous sentences (English (left) and Chinese (right)) from the Wikipedia data corpus in the figure for illustration. In each bar chart, the $i$-th element is the correlation calculated between the $i$-th word and the corresponding language embedding. Even though Chinese words are quite different from western languages, they both display the popularity of words to some degree.}
\label{fig:vis-correlation}
\end{center}
\end{figure}

\subsection{Revisiting Language Embedding}
We are interested in the role of the language embeddings in learning multilingual representations through Transformer layers. Assume that there are $N$ languages in the multilingual data corpus. Following the notations in Section 2.1, we denote $l_t \in \mathcal{R}^d$ as the language embedding for the $t$-th language, where $t\in[N]$. Denote $s=(w_1,....,w_n)$ as a sentence in the $t$-th language, where each $w_i\in \mathcal{R}^d$ is the word embedding. It is easy to show that when either applying the attaching approach or the additive approach, the self-attention (Eq.\ref{eqn:self-attn}) will calculate the correlation between word and language embeddings through dot-product. We show this explicitly for the additive approach while the analysis and results for the attaching approach are similar.

In the additive approach, the input\footnote{Usually, positional embedding is another term that will be added to the word embedding in the input. We omit the positional embedding for a better illustration of how language embeddings interact with words. The conclusions will not change if we take the positional embedding together into consideration. Recent works also show that positional embedding is not an essential 
term in the input, see \cite{ke2020rethinking,shaw2018self}.}  to the Transformer model is $\{w_i+l_t\}_{i=1}^{n}$. Then, in the self-attention module of the first Transformer layer, the correlation term $\alpha_{ij}$ in Eq.\ref{eqn:self-attn} can be expanded as:
\begin{align}\label{eqn:expand-attn}
    \alpha_{ij} &= \frac{((w_i+l_t)W^{Q,1}) ((w_j+l_t)W^{K,1})^T}{\sqrt{d}} \\
                      &= \frac{(w_iW^{Q,1}) (w_jW^{K,1})^T}{\sqrt{d}} + \frac{(l_tW^{Q,1}) (w_jW^{K,1})^T}{\sqrt{d}} \notag\\
                      &+ \frac{(w_iW^{Q,1}) (l_tW^{K,1})^T}{\sqrt{d}} + \frac{(l_tW^{Q,1}) (l_tW^{K,1})^T}{\sqrt{d}}. \notag
\end{align}
It can be seen that there are four terms in the expansion: word $\rightarrow$ word, language $\rightarrow$ word, word $\rightarrow$ language and language $\rightarrow$ language correlations. The first term characterizes the relationships between a pair of words, and the language embedding involves in the calculation of the other three terms. It is obvious to see that the last term is redundant since the value is a constant for every $i,j$. Adding the same constant to each dimension of the input will not change the output of the Softmax function. 

The two terms in the middle calculate the correlations between word embedding and language embedding. In particular, it can be seen from Eq.~\ref{eqn:expand-attn}, any word $w_i$ will calculate with the same language embedding $l_t$ in sentence $s$. We argue such correlations cannot reflect how two words in language $t$ correlate (e.g., have similar meanings) with each other. To study the function of the two terms, we download the officially released XLM15 model \cite{lample2019crosslingual}, and calculate those values on sampled sentences from the Wikipedia data corpus. We showcase the results in Figure \ref{fig:vis-correlation}. Empirically, we observe that the values seem to reflect the popularity of a word appearing in a language to a certain extent, e.g., ``the'' and ``of'' have relatively large dot-products with the language embedding of ``English''. But obviously, word-frequency correlation might not reflect the word-semantic correlation well.

\subsection{From Language Embedding to Language Projection}

The discussion above reveals some issues in the previous approaches and further motivates us to think about a better way to encode language information into a multilingual Transformer model. Note that the model uses a shared vocabulary, and a word-unit (or sub-word) may appear in multiple languages. We argue that for any language, the language encoding should provide the Transformer with language-specific meanings of the words in a sentence.  

Our idea is to use language projection instead of language embedding, which can provide language-specific word representations by \textit{projecting} the word embedding to language-specific semantic space. We denote $\phi_t: \mathcal{R}^{d} \rightarrow \mathcal{R}^{d}, t\in[N]$ as the projection function for the $t$-th language. For any sentence $s=(w_1,....,w_n)$ from language $t$ , we first project each word $w_i$ to $\phi_t(w_i)$ which characterizes the semantic meaning of $w_i$ in language $t$. After the projection, the language-specific word embedding $\phi_t(w_i)$ will be used as input to the Transformer model instead of $w_i$. Mathematically, we have the following form in the self-attention calculation:
\begin{equation}
    \label{eqn:language_function}
    \alpha_{ij} = \frac{(\phi_t(w_i)W^{Q,1})(\phi_t(w_j)W^{K,1})^T}{\sqrt{d}}
\end{equation}

We call our method Cross-lingual Language Projection (XLP for short). Linear projection is a popularly used semantic projection in many previous works \cite{vaswani2017attention, conneau2017word} and we use linear projection function in XLP. We define $\phi_t(w_i)=w_i P_t$, where $P_t$ is a $d\times d$ matrix. Then Eq.\ref{eqn:language_function} becomes

\begin{equation}
    \label{eqn:language_projection}
    \alpha_{ij} = \frac{((w_iP_t)W^{Q,1})((w_jP_t)W^{K,1})^T}{\sqrt{d}}
\end{equation}

It is worth noting that introducing $P_t$ can be viewed as a decoupling of the original projection matrices in the self-attention module. Essentially, $P_t$ learns language-specific projection to transform the word embedding to the language-specific semantic space. At the same time, $W^Q$ and $W^K$ still learn to project the semantic information to proper subspace as in the standard monolingual Transformer. 

In Figure \ref{fig:xlp-illustration}, we illustrate XLP and compare it with the language embedding (the additive approach). It can be seen that using the language embedding is equivalent to shifting the word embedding space by a language-specific bias. In contrast, our proposed language projection projects the word embedding to language-specific semantic space. Thus, the self-attention module can obtain the language-specific word correlations and learn more efficiently.

\subsection{Implementation and Discussions}

XLP is very easy to implement in any open-source deep learning framework like PyTorch \citep{paszke2019pytorch}. The overall pseudo code is shown in Algorithm \ref{alg:code}. More details are presented as follows.

\begin{algorithm}[t]
\caption{Pseudocode of XLP in a PyTorch-like style.}
\label{alg:code}

\definecolor{codeblue}{rgb}{0.25,0.5,0.5}
\lstset{
  backgroundcolor=\color{white},
  basicstyle=\fontsize{7.2pt}{7.2pt}\ttfamily\selectfont,
  columns=fullflexible,
  breaklines=true,
  captionpos=b,
  commentstyle=\fontsize{7.2pt}{7.2pt}\color{codeblue},
  keywordstyle=\fontsize{7.2pt}{7.2pt},
%  frame=tb,
}
\begin{lstlisting}[language=python]
# x: mini-batch with N samples
# positions: position index of each token
# lang_index: current language index
# language_projection: XLP function 
#        [nn.Linear(dim, dim) for _ in range(num_langs)]

# encode words to distributed representation
input = word_embedding(x)

# select corresponding XLP function by current language and project
input = language_projection[lang_index](input)

# add position embedding afterwards
input = input + position_embedding(positions)

# send to Transformer Module
output = TransformerEncoder(input)
\end{lstlisting}
\end{algorithm}
 
\textbf{Incorporate XLP with the positional embedding.} Positional encoding is an essential component in Transformer since other main components of the model are entirely invariant to sequence order. The absolute positional encoding is the most popularly used one, which provides each position an embedding vector. The positional embedding is added to the word embedding, which is found significantly helpful at learning the contextual representations of words at different positions. In XLP, the language projection is only applied to the word embedding. That being said, for any sentence, we first project the word embedding using XLP and then add it to the positional embedding. 

\textbf{The increase of parameters and efficiency.} The language embedding approaches use $N$ $d$-dimentional vectors where $N$ is the number of languages and $d$ is the embedding dimension. XLP uses $N$ $d\times d$ matrice, which is slightly larger than previous approaches. Taking XLM15 \citep{lample2019crosslingual} architecture as an example, the newly introduced language projection parameters are about 15M, which is only about 6\% of the 250M parameters in XLM15. Since XLP doesn't relate to the number of Transformer layers, the increase of parameters can be ignored when using deeper models. In terms of efficiency, XLP only needs an additional linear transformation in the input layer, which introduces barely computational overhead comparing to the stacked Transformer layers.

\section{Experiment}
We conduct extensive experiments on multilingual language understanding and multilingual machine translation tasks to verify our proposed XLP. The codes are implemented based on \texttt{fairseq}\footnote{\url{https://github.com/pytorch/fairseq/}} \citep{ott2019fairseq} and \texttt{XLM}\footnote{\url{https://github.com/facebookresearch/XLM/}} in \texttt{PyTorch} \citep{paszke2019pytorch}. Models are trained on 16/8 NVIDIA Tesla V100 GPUs with mixed-precision \citep{micikevicius2018mixed} for the multilingual language understanding / machine translation tasks respectively.

\subsection{Multilingual Language Understanding}

\begin{table*}[t]
\caption{\textbf{Results on XNLI cross-lingual natural language inference.} We report the accuracy on each language and the averaged accuracy. The languages are sorted by the resource magnitude order. \#lg is the number of languages used in the pre-training corpus.}
\label{tab:XNLI}
\vskip 0.1in
\begin{center}
\begin{small}
\renewcommand\tabcolsep{2.4pt}
\begin{tabular}{l|c|ccccccccccccccc|c}
\toprule
  & \textbf{\#lg} & \textbf{en} & \textbf{fr} & \textbf{es} & \textbf{de} & \textbf{el} & \textbf{bg} & \textbf{ru} & \textbf{tr} & \textbf{ar} & \textbf{vi} & \textbf{th} & \textbf{zh} & \textbf{hi} & \textbf{sw} & \textbf{ur} & \textbf{Avg} \\
\midrule
\multicolumn{17}{l}{\textit{Cross-Lingual Transfer (Fine-tune the multilingual model on English training set)}} \\
\midrule
\citet{conneau2017word} & - & 73.7 & 67.7 & 68.7 & 67.7 & 68.9 & 67.9 & 65.4 & 64.2 & 64.8 & 66.4 & 64.1 & 65.8 & 64.1 & 55.7 & 58.4 & 65.6 \\
\citet{artetxe2019massively} & - & 73.9 & 71.9 & 72.9 & 72.6 & 73.1 & 74.2 & 71.5 & 69.7 & 71.4 & 72.0 & 69.2 & 71.4 & 65.5 & 62.2 & 61.0 & 70.2 \\
\midrule
mBERT \citep{devlin2019bert}  & 102 & 82.1 & 73.8 & 74.3 & 71.1 & 66.4 & 68.9 & 69.0 & 61.6 & 64.9 & 69.5 & 55.8 & 69.3 & 60.0 & 50.4 & 58.0 & 66.3 \\
XLM100 \citep{lample2019crosslingual} & 100 & \textbf{83.7} & 76.2 & 76.6 & 73.7 & 72.4 & 73.0 & 72.1 & 68.1 & 68.4 & \textbf{72.0} & 68.2 & 71.5 & 64.5 & 58.0 & 62.4 & 71.3 \\
XLM15 \citep{lample2019crosslingual} & 15 & 83.2 & 76.5 & 76.3 & 74.2 & 73.1 & 74.0 & 73.1 & 67.8 & 68.5 & 71.2 & 69.2 & 71.9 & \textbf{65.7} & 64.6 & 63.4 & 71.5 \\
\textbf{XLP} & 15 & 83.2 & \textbf{76.6} & \textbf{78.3} & \textbf{74.5} & \textbf{75.5} & \textbf{74.7} & \textbf{73.4} & \textbf{70.0} & \textbf{72.3} & \textbf{72.0} & \textbf{69.3} & \textbf{73.8} & 65.2 & \textbf{67.1} & \textbf{64.9} & \textbf{72.7} \\
\midrule
\multicolumn{17}{l}{\textit{Translate-Train (Fine-tune the multilingual model on each training set)}} \\
\midrule
XLM100 \citep{lample2019crosslingual} & 100 & 82.9 & 77.6 & 77.9 & 77.9 & 77.1 & 75.7 & 75.5 & 72.6 & 71.2 & 75.8 & 73.1 & 76.2 & 70.4 & 66.5 & 62.4 & 74.2\\
\textbf{XLP} & 15 & \textbf{83.2} & \textbf{79.3} & \textbf{79.9} & \textbf{78.4} & \textbf{78.7} & \textbf{78.6} & \textbf{76.4} & \textbf{75.0} & \textbf{75.6} & \textbf{76.2} & \textbf{74.7} & \textbf{77.7} & \textbf{70.7} & \textbf{70.8} & \textbf{64.4} & \textbf{76.0}\\
\bottomrule
\end{tabular}
\end{small}
\end{center}
\vskip -0.1in
\end{table*}

\textbf{Model configurations.} For a fair comparison, we implement XLP based on XLM15 \citep{lample2019crosslingual} (250M parameters) architecture. Specifically, XLP consists of 12 Transformer encoder layers. For each layer, the dimensions of hidden representation and feed-forward layer are set to 1024 and 4096, respectively. The number of heads in the attention module is set to 8. XLM15 model supports 15 languages, and we use 15 projection matrices with shape 1024$\times$1024 in XLP. 

\textbf{Pre-training.} Following \citet{lample2019crosslingual}, we use Wikipedia corpus (15 languages) for pre-training. The detailed descriptions of the dataset can be found in Appendix A. We perform a couple of consecutive pre-processing steps following \citet{lample2019crosslingual}: normalizing, lower-casing, tokenizing the texts by Moses decoder\footnote{\url{https://github.com/moses-smt/mosesdecoder}} \citep{koehn-etal-2007-moses} and finally applying the byte pair encoding (BPE)\footnote{\url{https://github.com/glample/fastBPE}} \citep{sennrich2016neural} with the same codes (size 80k) and vocabulary (size 95k) from the XLM15. We use masked language modeling \citep{devlin2019bert} as the training objective. The model is trained for 750k steps. The batch size is set to 64 per GPU,  which is the same as the XLM \citep{lample2019crosslingual}. The gradient is accumulated every 4 optimization steps. Detailed description of the settings are presented in Appendix B.1.

\textbf{Fine-tuning.}  We use XNLI (\textbf{Cross}-lingual \textbf{N}atural \textbf{L}anguage \textbf{I}nference) \citep{conneau2018xnli} as the downstream evaluation benchmark to compare our proposed model with the baselines. Given a premise sentence and a hypothesis sentence in a language, the goal of the task is to predict whether the premise entails the hypothesis (entailment), contradicts the hypothesis (contradiction), or neither (neutral). XNLI dataset contains 15 languages, including low-resource languages such as Swahili and Urdu. 

Following \citet{lample2019crosslingual}, we evaluate the pre-trained models on the XNLI tasks in two settings. The first setting is called Cross-Lingual Transfer, in which we fine-tune the model on the English training set and evaluate it on the test sets of all languages. The second setting is called Translate-Train, in which we fine-tune and evaluate the model on the dataset of each language, respectively. For all the downstream experiments, we strictly follow \citet{lample2019crosslingual} for the hyperparameter configuration and search space using the official script\footnote{\url{https://github.com/facebookresearch/XLM/\#fine-tune-your-xlm-model-on-cross-lingual-classification-xnli}}.

\textbf{Results.}  The fine-tuning performance on XNLI is presented in Table \ref{tab:XNLI}. Languages are ordered according to the resource magnitude. We compare our model with five baselines: an LSTM-based model~\citep{conneau2018xnli}, a supervised model trained using translation pairs~\citep{artetxe2019massively}, the multilingual BERT model (mBERT)~\citep{devlin2019bert} and officially released XLM models (XLM15, XLM100). For all baselines, we use the number reported in their original papers.

It can be easily seen that XLP outperforms all baselines significantly. In the Cross-Lingual Transfer setting, XLP obtains 72.7\% averaged accuracy, which outperforms the XLM \citep{lample2019crosslingual} and mBERT \citep{devlin2019bert} by 1.2\% and 6.4\% respectively. When we fine-tune the pre-trained model on each language respectively (Translate-Train), XLP still achieves 1.8\% improvement over XLM. 

Moreover, XLP obtains a more balanced performance over the 15 languages. For high-resource languages, our approach is competitive (in English) or slightly better (in French, Greek)  than previous works. For extremely low-resource languages such as Swahili and Urdu, XLP outperforms XLM by a significant margin (2.5\% and 1.5\%). All the results indicate that our proposed XLP can help the model learn the multilingual sentence representation better.

\subsection{Multilingual Machine Translation}

\begin{table*}[t]
\caption{\textbf{BLEU scores of 6 languages $\rightarrow$ English on IWSLT dataset.} Language embedding approaches are compared as baselines. The attaching approach attaches the language embedding to the beginning of the sentence, while the additive approach adds the language embedding to the word embedding at each position.}
\label{tab:multi-trans}
\vskip 0.1in
\begin{center}
\begin{small}
\begin{tabular}{l|rrrrrr|r}
\toprule
\textbf{Model} & \textbf{ar-en} & \textbf{de-en} & \textbf{es-en} & \textbf{fr-en} & \textbf{tr-en} & \textbf{zh-en} & \textbf{Avg}\\
\cmidrule(lr){1-1} \cmidrule(lr){2-2} \cmidrule(lr){3-3} \cmidrule(lr){4-4} \cmidrule(lr){5-5} \cmidrule(lr){6-6} \cmidrule(lr){7-7} \cmidrule(lr){8-8}
Language Embedding (attaching) & 30.5 & 33.7 & 40.1 & 41.9 & 23.3 & 18.5 & 31.3 \\
Language Embedding (additive)  & 30.6 & 33.8 & 40.1 & 42.0 & 23.2 & 18.4 & 31.4 \\
\textbf{XLP}  & \textbf{31.0} & \textbf{34.4} & \textbf{40.6} & \textbf{42.2} & \textbf{24.2} & \textbf{18.7} & \textbf{31.9} \\
\bottomrule
\end{tabular}
\end{small}
\end{center}
\vskip -0.1in
\end{table*}

\begin{figure}[ht]
\centering
\vskip 0.1in
\begin{subfigure}{.23\textwidth}
  \centering
%   \vskip 0.1in
  \includegraphics[width=1.46in]{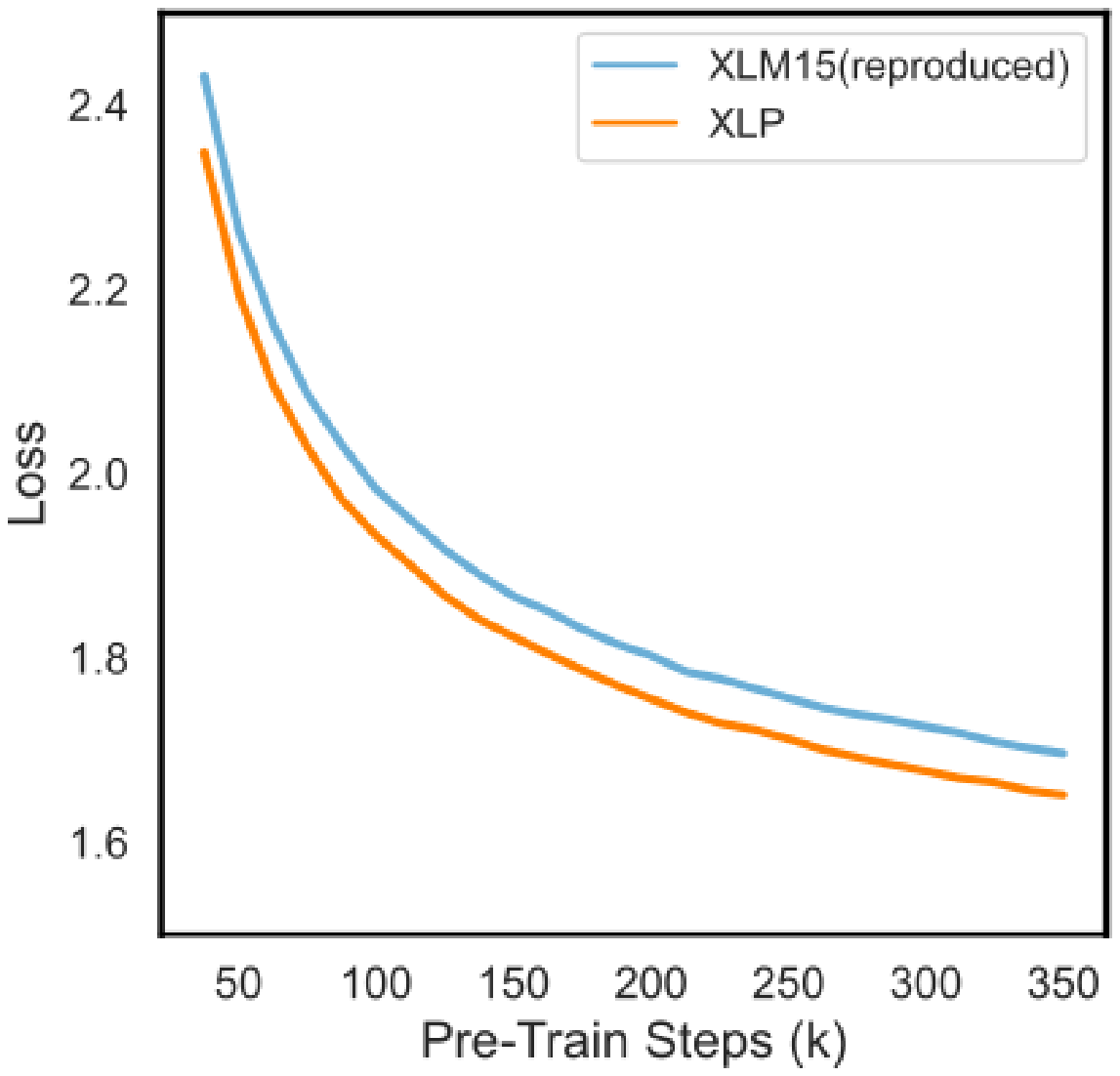} \vskip -0.01in 
  \caption{Valid loss (Pre-training).}
  \label{fig:pretrain-loss}
\end{subfigure}
\begin{subfigure}{.23\textwidth}
  \centering
  \includegraphics[width=1.50in]{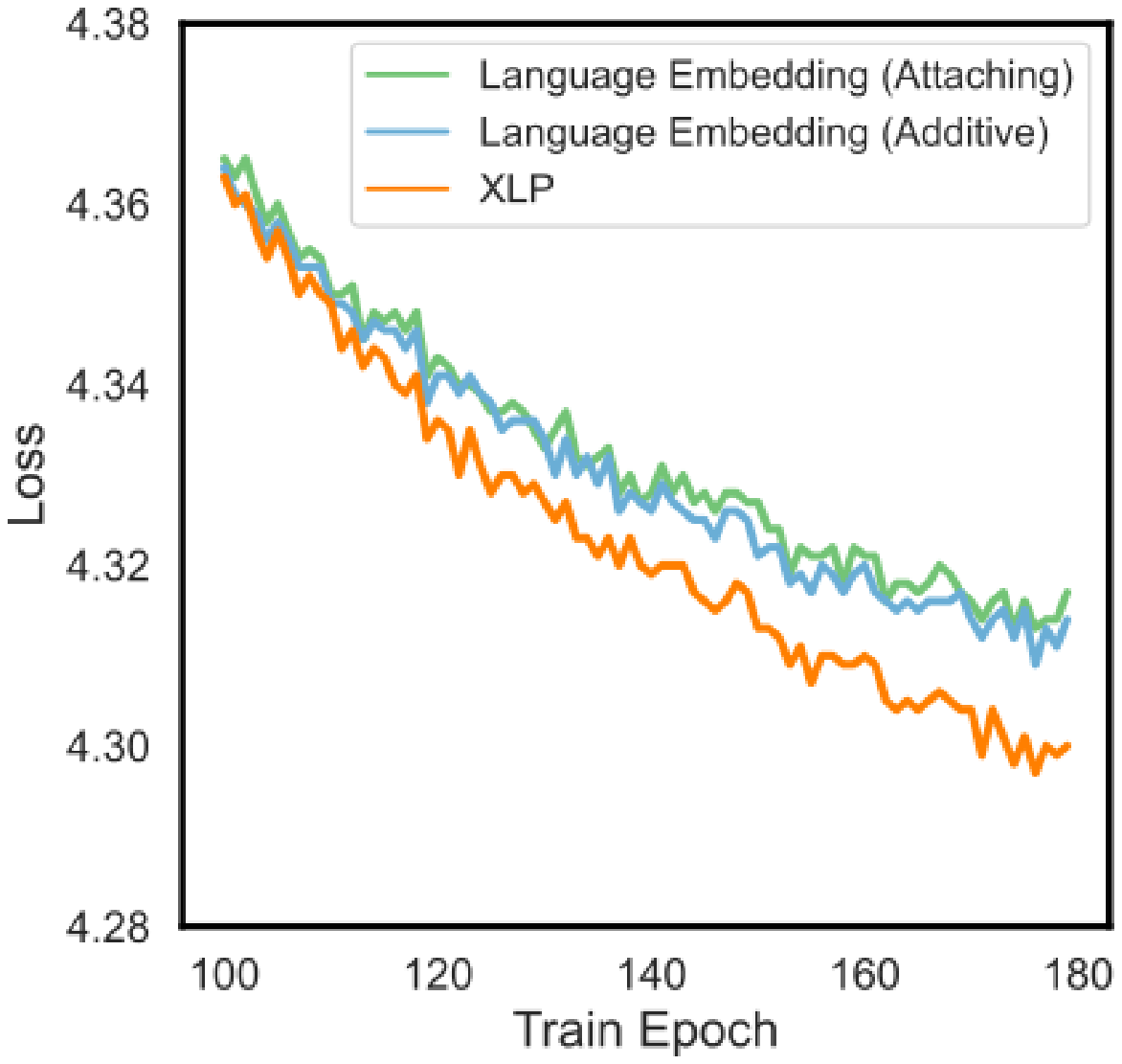}  
  \caption{Valid loss (Translation).}
  \label{fig:translate-loss}
\end{subfigure}
\caption{\textbf{Validation loss curve of different methods.} 
We plot the validation loss curve in both the cross-lingual language model pre-training and the multilingual translation task. It can be seen that XLP trains faster than previous approaches in both tasks.}
\label{fig:curve}
\vskip -0.1in
\end{figure}

\textbf{Model Configurations.} Following \citet{vaswani2017attention}, we use a 6-layer encoder-decoder-based Transformer in the machine translation tasks. The dimension of the hidden representation and the feed-forward layer is set to 512 and 1024 respectively. The number of the heads in the attention module is set to 4. We evaluate two language embedding approaches described in Section 2.2 as our baseline,  the attaching approach and the additive approach. All experiments use the same training and inference configurations. Detailed description of the settings are presented in Appendix B.2. 

\textbf{Datasets.} We collect 6 languages $\leftrightarrow$ English translation pairs from IWSLT evaluation campaign\footnote{\url{https://wit3.fbk.eu/2014-01}} (IWSLT 2014). The details about the datasets can be found in Appendix A. All the sentences are first tokenized with Moses tokenizer and then segmented into subword symbols using BPE. We learn the BPE merge operations across all the languages by setting the size of the BPE codes to 30000 and obtain a joint vocabulary with size 38413.

\textbf{Training and Inference.} We concatenate all the datasets to train a universal multilingual translation model. The mini-batch size is set to 4096 tokens per GPU. We use Adam \citep{kingma2017adam} as the optimizer, and set the hyperparameter $\epsilon$ to 1e-8 and ($\beta1,\beta2$) to (0.9, 0.98). The peak learning rate is set to 5e-4 with a 4k-step warm-up stage followed by an inverse square-root learning rate scheduler. We set the dropout probability to 0.3 and weight decay to 1e-4. Label smoothed cross-entropy is used as the objective function by setting $\epsilon = 0.1$ \citep{szegedy2015rethinking}. The number of training epoch is set to 180. We evaluate the translation quality by tokenized BLEU with sacreBLEU \footnote{\url{https://github.com/mjpost/sacrebleu}} \citep{post-2018-call}.

\textbf{Results.} The multilingual translation results are presented in Table \ref{tab:multi-trans}. First, our proposed XLP consistently outperforms the language embedding approaches (attaching or additive) and achieves an average 0.5 BLEU improvement comparing to the language embedding approaches. Besides, the experimental results also show that XLP works better on the low-resource \textit{TR} $\rightarrow$ \textit{EN} dataset (1.0 BLEU improvement). The overall comparison of the multilingual translation task also demonstrates the effectiveness of our proposed XLP.

\subsection{Discussion}

\textbf{Training Efficiency.}  By using the language projection, the Transformer model can receive concrete language-specific semantic information as input, which makes the model easier to train. To show this, we study the validation loss curves of XLP and the baselines. 

In cross-lingual language model pre-training, since XLM \citep{lample2019crosslingual} did not release any intermediate model checkpoints, we pre-trained both the XLM and the XLP using the same pre-training hyperparameters and check the intermediate model performance. In Figure \ref{fig:pretrain-loss}, we plot the pre-training validation loss of XLM and XLP for the first 350K steps. In multilingual machine translation, we compare the validation loss curve of XLP and previous language embedding approaches. The result is shown in Figure \ref{fig:translate-loss}. It can be easily seen that our proposed XLP 
can use fewer steps to achieve comparable validation loss than previous approaches.  

\begin{figure*}[ht]
\begin{subfigure}{.5\textwidth}  
  \centering  
  \includegraphics[width=2.7in]{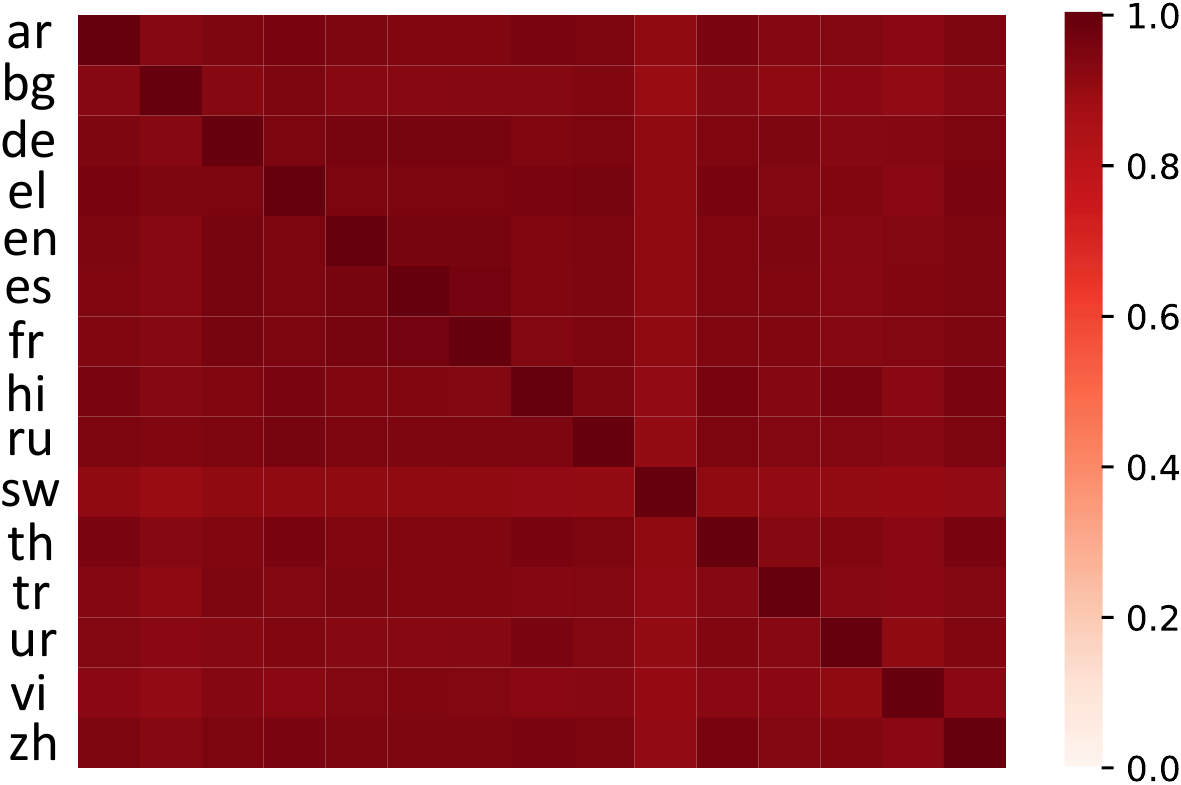}    
%   \vskip -0.1in  
  \caption{XLM}
  \label{fig:xlm-batch-vis}
\end{subfigure}
\begin{subfigure}{.5\textwidth}  
  \centering  
  \includegraphics[width=2.7in]{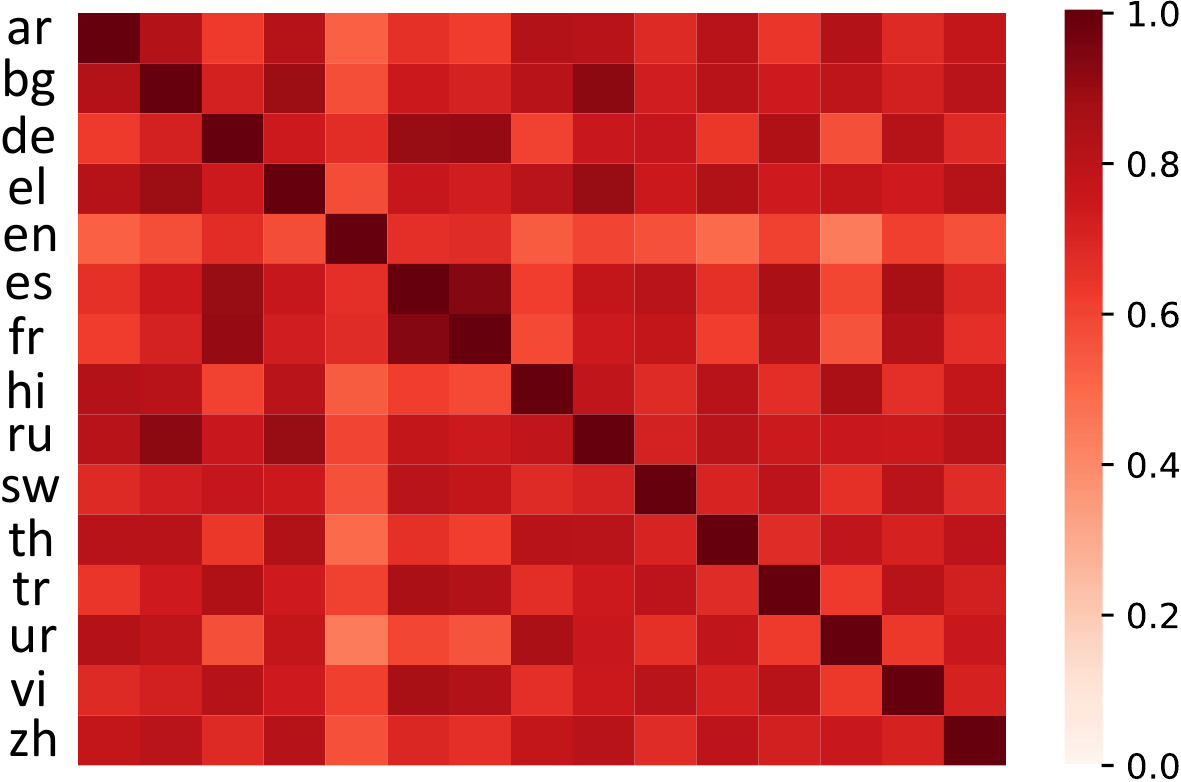}    
%   \vskip -0.1in  
  \caption{XLP}
  \label{fig:xlp-batch-vis}
\end{subfigure}
\caption{\textbf{Cosine similarity of processed words across different languages.} The word embeddings are processed using language embeddings in XLM (a) and XLP (b) }
\label{fig:vis-xlm-xlp}
\end{figure*}

\begin{figure*}
\begin{subfigure}{.5\textwidth}  
  \centering 
%   \vskip 0.15in
  \includegraphics[width=2.7in]{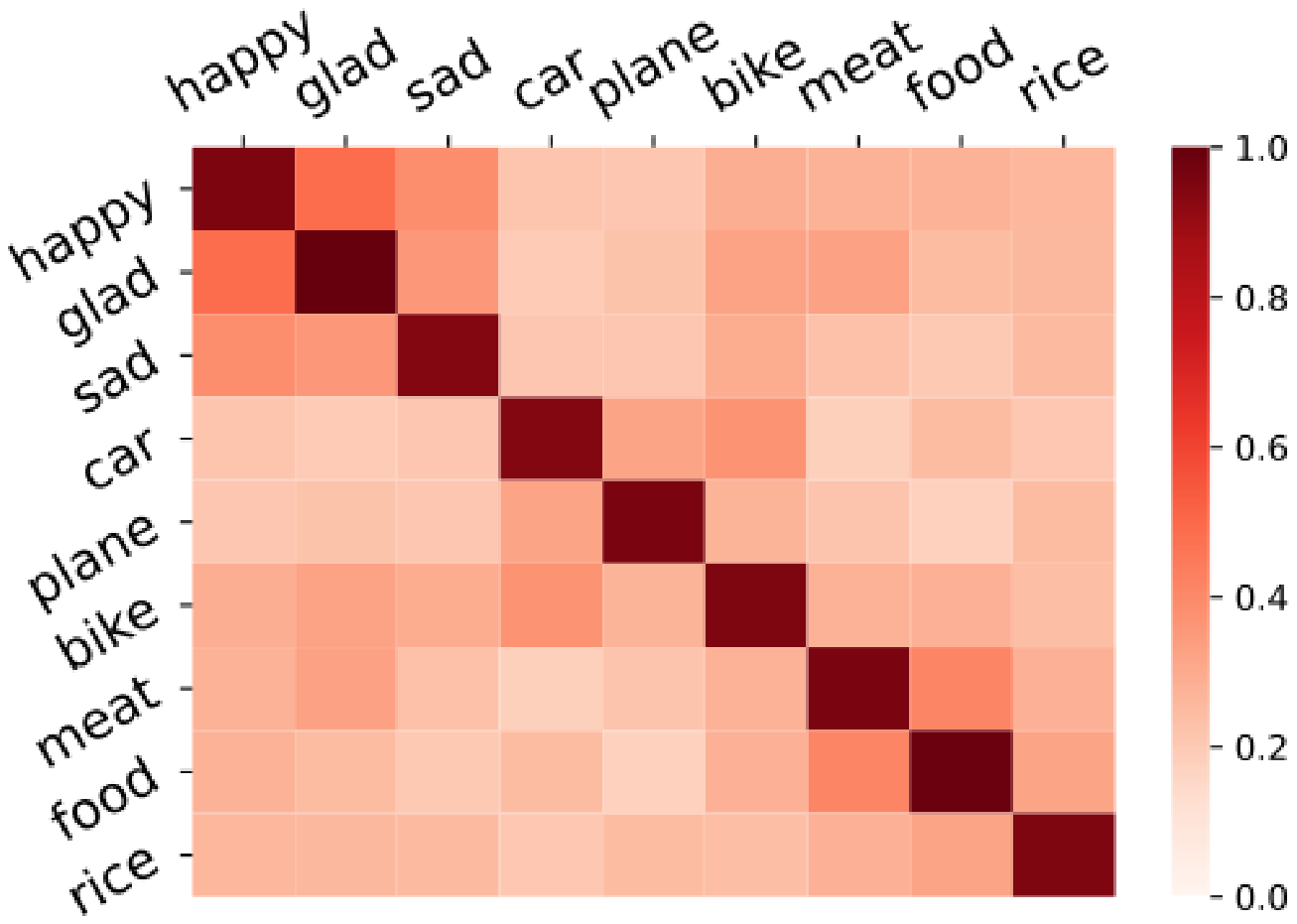}
%   \vskip -0.1in
  \caption{XLM}
  \label{fig:en-xlm-vis}
\end{subfigure}
\begin{subfigure}{.5\textwidth} 
  \centering  
%   \vskip 0.15in
  \includegraphics[width=2.7in]{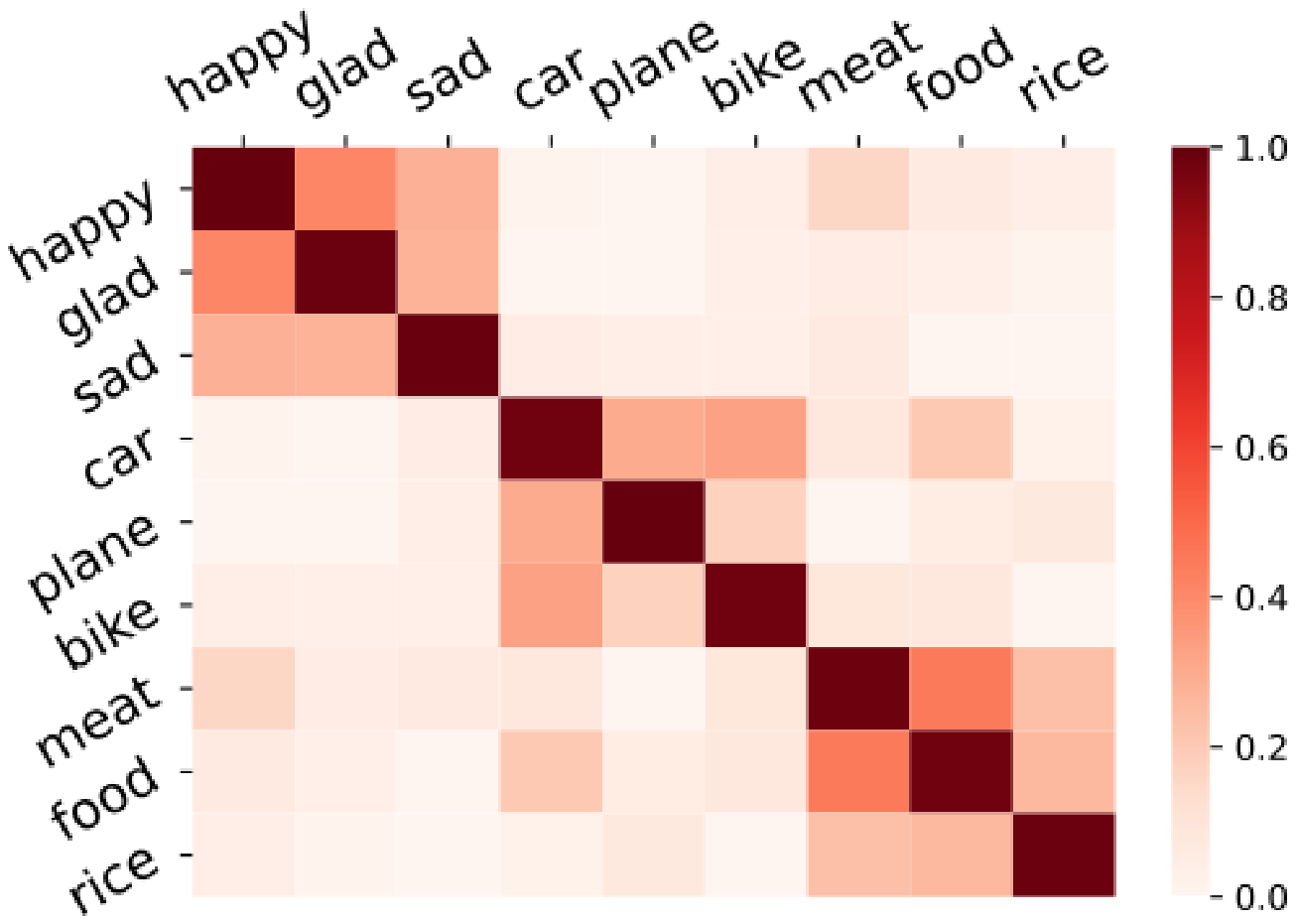}  
%   \vskip -0.1in
  \caption{XLP}
  \label{fig:en-xlp-vis}
\end{subfigure}
% \vskip -0.15in
\caption{\textbf{Cosine similarity of different processed words in ``English''.} The word embeddings are processed using language embeddings in XLM (a) and XLP (b).}
\label{fig:vis-xlm-xlpcorss-words}
\end{figure*}

\textbf{Cross-Lingual Transfer Gap.} The XTREME benchmark \citep{hu2020xtreme}  proposed to use the cross-lingual transfer gap to evaluate the multilingual models. The gap is measured by the difference between the performance on the English data and the averaged performance on all other language data. In such a way, we can estimate how powerful the multilingual model is when transferring its knowledge from English to other languages. As shown in Table \ref{table: xTansGap}, in terms of this metric, our proposed XLP significantly outperforms the mBERT, XLM100 and XLM15 by 5.7\%, 2.7\% and 1.3\% respectively on XNLI.

\begin{table}[t]
\caption{\textbf{Cross-lingual Transfer Gap Score.} The score is calculated as the difference between the performance on the English data and the averaged performance on all other language data in XNLI.}
\vskip 0.2in
\label{table: xTansGap}
\begin{center}
\begin{small}
% \begin{sc}
\begin{tabular}{lc}
\toprule
\textbf{Models} & \textbf{XNLI} \\
\midrule
mBERT \citep{devlin2019bert}  & 16.9 \\
XLM100 \citep{lample2019crosslingual}   & 13.9 \\
XLM15 \citep{lample2019crosslingual}   & 12.5 \\
XLP    & \textbf{11.2} \\
\bottomrule
\end{tabular}
% \end{sc}
\end{small}
\end{center}
\vskip -0.2in
\end{table}

\subsection{Visualization.} 
To better understand the learned language projections in XLP, we design methods to visualize them. Through the experiments,  we empirically verify that the language-specific projections indeed encode the language information and help the model efficiently capture the language-specific word correlations, while language embeddings do not.

First, we investigate whether the language encoding approaches learn language-specific information. The high-level idea is that if a language encoding approach learns languages specific information, the embeddings of a word processed by the language encoding approach should be different for different languages. To show this, we compare the released XLM15 model with our pre-trained XLP model for 15 languages. We first sample some English words (sub-words) from the multilingual vocabulary. For each word, we process the word embedding by different language encoding approaches (additive v.s. projection) and obtain language-specific embeddings of the same word. We then calculate the cosine similarity of these embeddings to see whether they are similar, which forms a $15\times15$ similarity matrix for each model.

As shown in Figure \ref{fig:xlm-batch-vis}, the elements in the similarity matrix of XLM are surprisingly high, while for XLP in Figure \ref{fig:xlp-batch-vis}, the element values are quite diverse. This demonstrates that our language projection indeed projects the word embeddings to different semantic spaces as we expect.

Taking one step further, we investigate whether the language encoding can help the model capture the word correlations. We select three topics first and then select three English words in each topic: [\textit{happy}, \textit{glad}, \textit{sad}], [\textit{car}, \textit{plane}, \textit{bike}] and [\textit{meat}, \textit{food}, \textit{rice}]. We process the word embeddings into ``English'' with different language encoding approaches (additive v.s. projection), and then compute the cosine similarity of the processed words. The word similarities are presented in Figure \ref{fig:en-xlm-vis} and \ref{fig:en-xlp-vis}. From the two figures, we can see that for XLP, the words in the same topic have strong correlations, while for words in different topics, the correlations are weak. However, in XLM, word pairs seem to have similar correlations. This suggests our proposed XLP captures the correlation of words better.

\section{Related Work}
Before the development of the Transformer model, Google built the first multilingual neural machine translation system (GMNMT) \citep{johnson2017googles} based on LSTM networks \citep{hochreiter1997long}, and introduced the language symbol with the attaching approach. Soon, this method is adopted to the Transformer-based multilingual translation system \citep{liu2020multilingual,tang2020multilingual}. XLM \citep{lample2019crosslingual} proposed to add the language embedding to word embedding at each position for multilingual language understanding tasks. \citet{huang2019unicoder,chi2019crosslingual} follow this language encoding approach further and develop more self-supervised objective functions.

There are several works studying the language-specific and language-agnostic parameters in a universal multilingual model. To build a multilingual machine translation system, early works aimed to increase the shared model capacity from the separated bilingual models to enhance the cross-lingual transfer \citep{dong-etal-2015-multi,zoph2016multisource,lee2017fully,firat2016multiway}. After the success of the first universal multilingual machine translation system \citep{wu2016googles,johnson2017googles}, emergent works \citep{blackwood2018multilingual,V_zquez_2019,escolano2020multilingual} started to investigate which components should be shared between languages and which components should be kept as language-specific in a universal multilingual model.

Recently, an interesting concurrent work \citep{zhang2021share} provided a comprehensive empirical study on the language-specific capacity in a universal multilingual machine translation model. This work suggests that using mixed language-specific and language-agnostic parameters in every sub-layers of the Transformer model is a better choice, letting the model learn to control the shared capacity by itself. However, this work does not take the language encodings into consideration but focuses on the upper Transformer layers. Being complementary to \citet{zhang2021share}, we investigate the previous language embedding approaches and propose to use the language-specific projections for better language encoding.

\section{Conclusion}
In this paper, we revisit the use of language embedding in the multilingual Transformer and identify several problems in the existing formulations. We propose a new approach called Cross-lingual Language Projection (XLP)  to address the issues. XLP uses language-specific transformations to project the word embeddings into language-specific semantic space, which achieves the purpose of appropriately encoding ``language'' in a multilingual Transformer. Extensive experiments demonstrate that the multilingual Transformer models using our proposed XLP consistently outperform those with previous language embedding approaches on multilingual language understanding and machine translation benchmarks. 

\bibliography{XLP}
\bibliographystyle{icml2021}

\newpage
\appendix

\section{Datasets}
\subsection{Pre-training}

Following \citet{lample2019crosslingual}, we use Wikipedia in 15 languages as the pre-training data corpus, whose size is roughly 42 GB. The dataset statistics are listed in Table \ref{table:wikidata}. 
We use WikiExtractor \footnote{\url{https://github.com/attardi/wikiextractor}} to extract raw sentences and perform a couple of consecutive pre-processing steps following \citet{lample2019crosslingual}: normalizing, lower-casing, tokenizing the texts by Moses decoder\footnote{\url{https://github.com/moses-smt/mosesdecoder}} \citep{koehn-etal-2007-moses} (Stanford Word Segmentor \footnote{\url{https://nlp.stanford.edu/software/segmenter.html}} for Chinese and PyThaiNLP \footnote{\url{https://github.com/PyThaiNLP/pythainlp}} for Thai) and finally apply the byte pair encoding (BPE)\footnote{\url{https://github.com/glample/fastBPE}} \citep{sennrich2016neural} with the same codes (size 80000) and vocabulary (size 95000) from the XLM15.

\begin{table}[h]
\caption{Wikipedia corpus for Pre-training.}
\label{table:wikidata}
\vskip 0.1in
\begin{center}
\begin{small}
\begin{tabular}{cccc}
\toprule
ISO code & Language & Language Samples (M) & Size (GiB)   \\
\cmidrule(lr){1-1} \cmidrule(lr){2-2} \cmidrule(lr){3-3} \cmidrule(lr){4-4}   
\textbf{ar} & Arabic & 4.69 & 1.1 \\
\textbf{bg} & Bulgarian & 1.60 & 0.34 \\
\textbf{de} & German & 18.85 & 5.4 \\
\textbf{el} & Greek & 1.41 & 0.37\\
\textbf{en} & English & 46.11 & 16.7 \\
\textbf{es} & Spanish & 12.29 & 3.2 \\
\textbf{fr} & French & 16.94 & 4.5  \\
\textbf{hi} & Hindi & 0.71 & 0.16 \\
\textbf{ru} & Russian & 13.81 & 3.8 \\
\textbf{sw} & Swahili & 0.23 & 0.03 \\
\textbf{th} & Thai & 0.87 & 0.28\\
\textbf{tr} & Turkish & 1.97 & 0.55\\
\textbf{ur)} & Urdu & 0.60 & 0.14\\
\textbf{vi} & Vietnamese & 4.22 & 0.72\\
\textbf{zh} & Chinese & 6.08 & 1.9\\
\bottomrule
\end{tabular}
\end{small}
\end{center}
\vskip -0.1in
\end{table}

\subsection{Fine-tuning}

XNLI (\textbf{Cross}-lingual \textbf{N}atural \textbf{L}anguage \textbf{I}nference) benchmark \citep{conneau2018xnli} is a cross-lingual extension of the NLI task. Given a premise sentence and a hypothesis sentence in a language, the goal of the task is to predict whether the premise entails the hypothesis (entailment), contradicts the hypothesis (contradiction), or neither (neutral). The XNLI dataset is constructed by extending the development and test sets of
the Multi-Genre Natural Language Inference Corpus (MultiNLI) to 15 languages, making 112,500 annotated pairs in total. For each language, we have 2490 samples for validation and 5010 samples for test.

Following \citet{lample2019crosslingual}, we evaluate the pre-trained models on the XNLI tasks in two settings. The first setting is called Cross-Lingual Transfer, in which we fine-tune the model on the English training set and evaluate it on the test sets of all languages. The second setting is called Translate-Train, in which we fine-tune and evaluate the model on the dataset of each language, respectively.

\subsection{Multilingual Machine Translation}

We collect 6 languages $\rightarrow$ English translation pairs from the IWSLT evaluation campaign in year 2014\footnote{\url{https://wit3.fbk.eu/2014-01}}. The dataset statistics are presented in Table \ref{table:iwslt}.

For pre-processing, all the sentences are first tokenized with Moses tokenizer and then segmented into subword symbols using Byte Pair Encoding (BPE). Note that we use the Stanford Word Segmenter to tokenize the sentences in Chinese. We learn the BPE merge operations across all the languages by setting the size of the BPE codes to 30000 and obtain a joint vocabulary with size 38413.

\begin{table}[h]
\caption{IWSLT datasets for Translation.}
\label{table:iwslt}
\vskip 0.1in
\begin{center}
\begin{small}
\begin{tabular}{ccccc}
\toprule
ISO Code & Language & Train & Valid & Test \\
\cmidrule(lr){1-1} \cmidrule(lr){2-2} \cmidrule(lr){3-5}
\textbf{ar} & Arabic & 139747 & 6352 & 5357 \\
\textbf{de} & German & 160239 & 7283 & 6750 \\
\textbf{es} & Spanish & 169030 & 7683 & 5593 \\
\textbf{fr} & French & 168152 & 7643 & 4493 \\
\textbf{tr} & Turkish & 109892 & 4995 & 5433 \\
\textbf{zh} & Chinese & 155615 & 7073 & 5099 \\
\bottomrule
\end{tabular}
\end{small}
\end{center}
\vskip -0.1in
\end{table}

\section{Training Configurations}
\subsection{Multilingual Language Understanding}
\paragraph{Baselines.}

We use several competitive baselines for comparison: (1) \citet{conneau2018xnli}: the baseline approach from the XNLI benchmark which is based on the LSTM \citep{hochreiter1997long}; (2) \citet{artetxe2019massively}: a supervised approach which uses 223 million parallel sentences; (3) mBERT \citep{devlin2019bert}: the multilingual BERT which is pre-trained with masked language modeling (MLM) on Wikipedia in 102 languages; (4) XLM \citep{lample2019crosslingual}: the MLM pre-trained multilingual model which uses language embedding to encode language information. XLM15 (250M parameters) is pre-trained on Wikipedia in 15 languages, while XLM100 (570M parameters) is pre-trained on Wikipedia in 100 languages.

\begin{table}[h]
\caption{Hyperparameters used in the Multilingual Language Understading experiments}
\label{table:understand_param}
\vskip 0.1in
\begin{center}
\begin{small}
\begin{tabular}{lcc}
\toprule
 & \textbf{Pre-training} & \textbf{Fine-tuning} \\
\midrule
\textit{Model Configuration} \\
\midrule
\textbf{Layers} & \multicolumn{2}{c}{12} \\
\textbf{Hidden Representation} & \multicolumn{2}{c}{1024} \\
\textbf{Feed-Forward Layer} & \multicolumn{2}{c}{4096} \\
\textbf{Heads} & \multicolumn{2}{c}{8} \\
\midrule
\textit{Hyperparameters} \\
\midrule
\textbf{Max Steps} & 750k & - \\
\textbf{Learning Rate} & 1e-4 & (1$\sim$8)e-5 \\
\textbf{Warm-up Steps} & 16k & 4k  \\
\textbf{Learning Rate Decay} & Inverse Sqrt & Inverse Sqrt \\
\textbf{Batch Size} & 64 & 16 \\
\textbf{Sequence Length} & 256 & 256 \\
\textbf{Adam $\epsilon$} & 1e-8 & 1e-8 \\
\textbf{Adam($\beta_1$, $\beta_2$)} & (0.9, 0.98) & (0.9, 0.999) \\
\textbf{Clip Norm} & 5 & 5 \\
\textbf{Dropout} & 0.1 & 0 \\
\textbf{Gradient Accumulation} & 4 & 1 \\
\bottomrule
\end{tabular}
\end{small}
\end{center}
\vskip -0.1in
\end{table}

\paragraph{Model Configurations and Training Details.}
The overall settings are summarized in Table \ref{table:understand_param}. To compare with XLM15, we build our model as a 12-layer Transformer. For each layer, the dimensions of hidden representation and feed-forward layer are set to 1024 and 4096, respectively. The number of heads in the attention module is set to 8. XLM15 model supports 15 languages, and we use 15 projection matrices with shape 1024$\times$1024 in XLP. 

We use masked language modeling as the objective of pre-training. We train the model for 750k steps. The batch size is set to 64 per GPU. Due to the limit of the GPU memory, we accumulate gradients every 4 optimization steps. Models are trained on 16 NVIDIA Tesla V100 GPUs with mixed-precision. Thus, the effective batch size is 4096, which is the same as the XLM15. The maximum sequence length is 256. The masked probability is set to 0.15, with replacing 80\% of the masked positions by {\tt[MASK]}, 10\% by randomly sampled words, and keep the remaining 10\% unchanged. We use Adam \citep{kingma2017adam} as the optimizer, and set the hyperparameter $\epsilon$ to 1e-8 and ($\beta1,\beta2$) to (0.9, 0.98). The peak learning rate is set to 1e-4 with a 16k-step warm-up stage followed by an inverse square-root learning rate scheduler. The dropout probability and the weight decay parameter are set to 0.1 and 1e-4 respectively.

During fine-tuning for the XNLI task, We search the learning rates (from 1e-6 to 8e-6) and batch size (8 or 16). For XLP, we fix the language-specific projection weights during fine-tuning. We use two ways to evaluate the pre-trained models: Cross-Lingual Transfer and Translate-Train, as described in Appendix A.2.

\subsection{Multilingual Machine Translation}
\paragraph{Baselines.}

We use two proposed baselines for comparison: (1) Language Embedding (attaching): we encode the language information by attaching a language-specific token at the beginning of the sentences, as stated in Section 2.2. (2) Language Embedding (additive): we encode the language information by adding the word embedding at each position with the language embedding, as stated in Section 2.2.

\begin{table}[h]
\caption{Hyperparameters used in the Multilingual Machine Translation experiments}
\label{table:translation_param}
\vskip 0.1in
\begin{center}
\begin{small}
\begin{tabular}{lc}
\toprule
 & \textbf{Translation} \\
\midrule
\textit{Model Configuration} \\
\midrule
\textbf{Layers for Encoder / Decoder} & 6 / 6\\
\textbf{Hidden Representation} & 512\\
\textbf{Feed-Forward Layer} & 1024 \\
\textbf{Heads} & 4\\
\midrule
\textit{Hyperparameters} \\
\midrule
\textbf{Training Iterations} & 180 epoch \\
\textbf{Learning Rate} & 5e-4 \\
\textbf{Warm-up Step} & 4k \\
\textbf{Learning Rate Decay} & Inverse Sqrt \\
\textbf{Tokens per batch} & 4096 \\
\textbf{Adam $\epsilon$} & 1e-8 \\
\textbf{Adam($\beta_1$, $\beta_2$)} & (0.9, 0.98) \\
\textbf{Dropout} & 0.3 \\
\textbf{Label Smooth $\epsilon$} & 0.1 \\
\textbf{Beam Search Size} & 5 \\
\textbf{Length Penalty} & 1.2 \\
\bottomrule
\end{tabular}
\end{small}
\end{center}
\vskip -0.1in
\end{table}

\paragraph{Model Configurations and Hyperparameters.}
The overall settings are summarized in Table \ref{table:translation_param}. Following \citet{vaswani2017attention}, we use a 6-layer encoder-decoder-based Transformer in the machine translation tasks. The dimension of the hidden representation and the feed-forward layer is set to 512 and 1024 respectively. The number of the heads in the attention module is set to 4.

For the multilingual model training, the mini-batch size is set to 4096 tokens per GPU. We use Adam \citep{kingma2017adam} as the optimizer, and set the hyperparameter $\epsilon$ to 1e-8 and ($\beta1,\beta2$) to (0.9, 0.98). The peak learning rate is set to 5e-4 with a 4k-step warm-up stage followed by an inverse square-root learning rate scheduler. We set the dropout probability to 0.3, and weight decay to 1e-4. Label smoothed cross entropy is used as the objective function by setting $\epsilon = 0.1$ \citep{szegedy2015rethinking}. The total training epoch is set to 180. During inference, we decode with beam search and set the beam size to 5 for all the languages. Length penalty is set to 1.2. We evaluate the translation quality by tokenized BLEU with sacreBLEU \footnote{\url{https://github.com/mjpost/sacrebleu}} \citep{post-2018-call}.

\end{document}